\documentclass{article}

\usepackage{microtype}
\usepackage{graphicx}
\usepackage{booktabs} % for professional tables

\usepackage{hyperref}

\usepackage[accepted]{icml2025}

\usepackage{amsmath}
\usepackage{amssymb}
\usepackage{mathtools}
\usepackage{amsthm}

\usepackage[capitalize,noabbrev]{cleveref}

\usepackage{pgfplots}
\pgfplotsset{compat=1.18} % 设置pgfplots兼容版本
\usepackage{multirow}
\usepackage{colortbl}
\usepackage{color}
\usepackage{subcaption} % 提供 subfigure 环境

\theoremstyle{plain}
\newtheorem{theorem}{Theorem}[section]

\theoremstyle{definition}
\newtheorem{definition}[theorem]{Definition}

\theoremstyle{remark}

\usepackage[textsize=tiny]{todonotes}

\icmltitlerunning{Improving Value-based Process Verifier via Structural Prior Injection}

\begin{document}

\twocolumn[
\icmltitle{Improving Value-based Process Verifier via Structural Prior Injection}



\icmlsetsymbol{equal}{*}

\begin{icmlauthorlist}
\icmlauthor{Zetian sun}{sch}
\icmlauthor{Dongfang Li}{sch}
\icmlauthor{Baotian Hu}{sch}
\icmlauthor{Jun Yu}{sch}
\icmlauthor{Min Zhang}{sch}
\end{icmlauthorlist}

\icmlaffiliation{sch}{Harbin Institute of Technology (Shenzhen), Shenzhen, China}

\icmlcorrespondingauthor{Zetian Sun}{zetiansun.cs@gmail.com}
\icmlcorrespondingauthor{Baotian Hu}{hubaotian@hit.edu.cn}

\icmlkeywords{Machine Learning, ICML}

\vskip 0.3in
]



\printAffiliationsAndNotice{}  % leave blank if no need to mention equal contribution

\begin{abstract}
In the Large Language Model(LLM) reasoning scenario, people often estimate state value via Monte Carlo sampling. Though Monte Carlo estimation is an elegant method with less inductive bias, noise and errors are inevitably introduced due to the limited sampling. To handle the problem, we inject the structural prior into the value representation and transfer the scalar value into the expectation of a pre-defined categorical distribution, representing the noise and errors from a distribution perspective. Specifically, by treating the result of Monte Carlo sampling as a single sample from the prior ground-truth Binomial distribution, we quantify the sampling error as the mismatch between posterior estimated distribution and ground-truth distribution, which is thus optimized via distribution selection optimization. We test the performance of value-based process verifiers on Best-of-N task and Beam search task. Compared with the scalar value representation, we show that reasonable structural prior injection induced by different objective functions or optimization methods can improve the performance of value-based process verifiers for about 1$\sim$2 points at little-to-no cost. We also show that under different structural prior, the verifiers' performances vary greatly despite having the same optimal solution, indicating the importance of reasonable structural prior injection. 

\end{abstract}

\section{Introduction}
The process verifier plays a crucial role in Large Language Model(LLM) reasoning scenario, which requires numerous intermediate steps and the quality of the intermediate reasoning step is pivotal\citep{DBLP:conf/iclr/LightmanKBEBLLS24}. Compared with outcome verifiers solely focusing on outcome correctness\citep{DBLP:journals/corr/abs-2110-14168}, process verifiers provide more detailed signals throughout the multi-step decision-making process including the final outcome, which can be an elegant replacement to outcome verifiers at little-to-no cost\citep{DBLP:conf/iclr/LightmanKBEBLLS24}. In the LLM reasoning scenario, only the outcome binary reward is accurate and has no inductive bias. We thus focus on the value-based process verifier, where the process signal is the value of the current state\citep{lu2024autopsv, DBLP:journals/corr/abs-2406-06592,DBLP:journals/corr/abs-2408-03314}. The value-based process verifier incorporates limited inductive bias: the Markov Decision Environment has only the outcome binary reward, and the discount factor is equal to 1. In such an environment, the value signal can be effectively annotated through Monte Carlo Sampling, as introduced in \citet{DBLP:conf/acl/WangLSXDLCWS24}.

Though Monte Carlo estimation is an unbiased estimation of the ground-truth value(i.e. the success rate of the current state to outcome), it's still inaccurate especially when the sampling quantity is limited. The first problem of the limited sampling quantity is that it can only represent limited different values. As the Monte Carlo estimation method tries to estimate the state value in a discrete sampling space, the differences in representation granularity between the estimated value and the ground-truth value will be unacceptable. The second problem of the limited sampling quantity is that the variance of Monte Carlo estimation is related to the ground-truth value $p$. When $p$ is close to a marginal value, the variance of the Monte Carlo estimation result is small and the estimation can be accurate. When $p$ is close to the middle value, the estimation will be inaccurate except for increasing the sampling quantity.

To address the first problem, people often use the mean-square error objective function and map the discrete estimated value into the contiguous space\citep{lu2024autopsv, DBLP:journals/corr/abs-2406-06592,DBLP:journals/corr/abs-2408-03314}. The mean-square error objective function introduces the distance prior and shows that the distance between two different values can be measured by the square of the difference between the two values, which extends the discrete estimated values into the contiguous space. To address the second problem, people often use the temporal difference method to update the value of the current state using values of subsequent states iteratively, greatly reducing the sampling variance at the cost of a larger sampling quantity\citep{DBLP:journals/tac/TsitsiklisR97,DBLP:journals/corr/abs-2407-04811}. However, in the LLM reasoning scenario, it's expensive to extend the sampling quantity as each sample requires a completed rollout to the final state, which limits the application of Monte Carlo estimation as the annotation of the state value.

In this paper, we propose the structural prior injection method. To clarify, motivated by the distance prior introduced by the mean-square error objective function that is used to bridge the gap between discrete labels and contiguous target space, we inject the structural prior into the na\"ive scalar Monte Carlo state value expression, representing the scalar value as the expectation of some pre-defined categorical distribution. Under a suitable prior definition, the Monte Carlo estimation result can be treated as a one-step sampling of the ground-truth target distribution that is derived from the ground-truth state value $p$. The transformation allows us to interpret the Monte Carlo sampling error and the coarse-grained signal as the mismatch between the ground-truth target distribution and the posterior distribution conditioned on a sample taken from the ground-truth target distribution, and the error is derived from the difficulty of recovering the ground-truth distribution given limited sampling data.

To handle the problem, we propose the distance metric called Statistics-based Distance, which is used to measure the distance between two categorical distributions conditioned on the probability of sampling the corresponding category from the ground-truth target distribution. The distance metric can be used to judge the reasonableness of the prior structural information to be injected, which can therefore guide the optimization approach. We optimize the categorical distribution via different objective functions, including the mean-square error and the cross-entropy(HL) loss\citep{DBLP:conf/icml/ImaniW18}, showing that the reasonable structural prior can improve the performance of value-based process verifiers on different objective functions and different tasks for about 1$\sim$2 points at little-to-no cost. Through ablation studies, we also show that different structural prior can result in very different performances.

Finally, we summarize our contributions in this paper:
\begin{itemize}
    \item We propose the structural prior injection method, transferring the Monte Carlo sampling error into the distribution mismatch problem, showing that the error is due to the difficulty of recovering the ground-truth distribution using limited sampling data.
    \item We propose the Statictics-based Distance metric, which can guide the posterior distribution selection and handle the mismatch problem and is supported by the experiments.
    \item We show that the improvements derive from the reasonable structural prior injection, which is a promising and meaningful direction for future work.
\end{itemize}
\section{Preliminaries}
We first formally review the concepts and formulations of the Markov decision process(MDP) and the Bellman Equation in the Large Language Model(LLM) reasoning scenario. Then, we review the Monte Carlo estimation method, which is used to estimate the state action value. Finally, we introduce the value-based process verifier.

\subsection{Markov decision process and Bellman Equation}\label{sec:mdp}
The Markov decision process can be represented as a 4-tuple $(\mathcal{S}, \mathcal{A}, \mathcal{P}, \mathcal{R})$, where $\mathcal{S}=\{s\}$ is the state space, $\mathcal{A}=\{a\}$ is the action space, $\mathcal{P}=\mathcal{P}(s_t,a_t,s_{t+1})$ is the transition probability measuring the probability that action $a_t$ in state $s_t$ at time t will lead to state $s_{t+1}$ at time t+1. In the deterministic scenario, the transition probability will be restricted to 1, showing that given $s_t$ and $a_{t}$, the next state will be $s_{t+1}$. $\mathcal{R}=\mathcal{R}(s_t,a_t,s_{t+1})$ is the reward function, measuring the reward received after transitioning from state $s_t$ to $s_{t+1}$ due to action $a_t$. The policy $\pi$ is a mapping from the state space $\mathcal{S}$ to the action space $\mathcal{A}$. A completed MDP consists of multiple states $\{s_1,s_2,...,s_t\}$, where each state $s_i$ first samples the next action $a_i$ using the policy $\pi$, then samples the next state $s_{i+1}$ according to the transition probability distribution given the current state $s_i$ and action $a_i$.

The Bellman Equation defines the relationship between neighboring values in an MDP. Given the state value $V$ defined as 
\begin{equation}\label{pre:value_def}
    V^{\pi}(s)=\mathbb{E}_{\pi}[G_t|S_t=s]
\end{equation}
where $\pi$ being the current policy and $s$ being the current state, and return $G_t$ defined as
\begin{equation}
    G_t=\sum_{k=0}^{\infty}\gamma \mathcal{R}_{t+k}
\end{equation}
where $\mathcal{R}_{t+k}$ is the abbreviation of $\mathcal{R}(s_{t+k},a_{t+k},s_{t+k+1})$ and $\gamma$ being the discount factor measuring the rate of reward decay, the expectation version of Bellman Equation is defined as
\begin{equation}\label{pre:bellman_raw}
\begin{aligned}
    V^{\pi}(s)&=\mathbb{E}_{\pi}[R_t|S_t=s]+\gamma\sum_{s'\in \mathcal{S}}P(s'|s)V^\pi(s') \\
    &=r(s)+\gamma\sum_{s'\in \mathcal{S}}P(s'|s)V^\pi(s')
\end{aligned}
\end{equation}
where $r(s)$ is defined as the expectation of immediate reward at state $s$.

In the LLM reasoning scenario, a completed reasoning path is split by a rule-based splitter like "$\backslash$n", forming a few non-overlapping reasoning steps. The action is defined as a single reasoning step, and the state is defined as the cumulative reasoning steps. Given current state $s_t$ and action $a_t$, the next state $s_{t+1}$ is uniquely determined, which shows that the MDP is deterministic. Another popular setting is to let the discount factor $\gamma$ equal one and the immediate reward equal zero, except for the last outcome reward being binary. Given a trajectory that has N steps, the MDP settings can be described as follows:
\begin{equation}\label{pre:reward_outcome}
    r(s_t)=\left\{
    \begin{aligned}
        0 & & {t<N} \\
        r & & {t=N}
    \end{aligned}
    \right.
\end{equation}
where $r$ is the outcome reward, $r=1$ if the outcome is correct, $r=0$ if the outcome is incorrect. Given the settings above, the Bellman Expectation Equation (i.e. Eqn.\ref{pre:bellman_raw}) can be simplified as follows:
\begin{equation}\label{pre:value_neigh}
    V^\pi(s)=\sum_{a\in\mathcal{A}}\pi(a|s)V^\pi(s')
\end{equation}
where $s'=s\bigcup\{a\}$ is the next state of $s$ given current state $s$ and action $a$.

\subsection{Monte Carlo Estimation}
Monte Carlo estimation is the method that estimates state value function given policy $\pi$ using Monte Carlo sampling in MDP. Rewriting Eqn.\ref{pre:value_def}, we have:
\begin{equation}
    V^\pi(s)=\mathbb{E}_\pi[G_t|S_t=s]\approx\frac{1}{N}\sum_{i=1}^N [G_t^{(i)}|S_t=s]
\end{equation}
where each $G_t^{(i)}$ is the return of a completed trajectory sampled from trajectories that $S_t=s$ using policy $\pi$. Monte Carlo estimation is an unbiased and effective estimation method. 

In the LLM reasoning scenario, the return $G_t$ can be simplified as:
\begin{equation}
    G_t=\sum_{k=0}^{\infty}\gamma \mathcal{R}_{t+k}=r
\end{equation}
where $r$ is defined after Eqn.\ref{pre:reward_outcome}. Then, the Monte Carlo estimation of state value $V$ can be simplified as:
\begin{equation}
    V^\pi(s)\approx\frac{1}{N}\sum_{i=1}^N [G_t^{(i)}|S_t=s]=\frac{1}{N}\sum_{i=1}^N [r^{(i)}|S_t=s]
\end{equation}
which shows that we can estimate the intermediate state value using the outcome reward.

\subsection{Value-based Process Verifier}
The value-based process verifier aims at estimating the state value during the reasoning process. Given the initial question $q$ and previous reasoning steps $\{a_1,a_2,...,a_t\}$, the value-based process verifier is asked to generate a scalar value that represents the state value of the current state $s_t=\{a_1,a_2,...,a_t\}$. Under the environment settings shown in \S\ref{sec:mdp}, given the ground truth answer, it's able to perform an automatic annotation approach to obtain process state value signals using the Monte Carlo method, and the state value is interpreted as the success rate of the outcome starting from the current state. We can thus formulate the value-base process verifier as a mapping from binary function to probabilities using parameter $\theta$:
\begin{equation}
    f_\theta(q, s_t)\rightarrow[0,1]
\end{equation}

\section{Methodology}

\subsection{Regression: from Scalar to Categorical Distribution}\label{sec:regression}
As we perform $k$ individual rollouts from the current state to calculate the Monte Carlo return and estimate the state value, each of the rollouts will return a binary value which is the outcome reward. There are totally $k+1$ different kinds of estimated state values, evenly ranging from 0 to 1 with step size being $\frac{1}{k}$. We can thus construct a classification objective and train a classifier to handle the problem. To clarify, we map the $k+1$ different estimated values into $k+1$ different bins, each bin represents one of the estimated values. Each time, given the question $q$ and the current state $s_t$, the value-based process verifier, which is a classifier, chooses one of the $k+1$ different bins and maps the selected bin to the value. We can optimize the value-based process verifier through cross-entropy loss as follows:
\begin{equation}
    \min_\theta\sum_{i=1}^N\log f_\theta(b_i|q,s_t),
\end{equation}
where $f_\theta$ is the function mapping from state space $\{q\}\bigcup s_t\in\mathcal{S}$ to one of the bins. $b_i\in\mathbb{N}^+$ is the bin index of estimated state value ranging from 1 to $k+1$. The $i$-th bin will then be mapped to the state values, for example, $\frac{i-1}{k}$. In such case, the predicted state value $f_\theta(\cdot|q,s_t)$ is defined in the discrete space while $V^\pi(s)$ is defined in the contiguous space, which poses a mismatch between the objective function and target value. 

In order to bridge the gap between the discrete estimated state value space and the contiguous target value space, researchers treat the value prediction problem as a regression task that includes the \textbf{distance prior}. Specifically, for two estimated values $\frac{i}{k}$ and $\frac{j}{k}$, the distance between them is the square of the difference. The distance prior allows us to optimize the discrete estimated state value in the contiguous value space, ensuring the predicted value makes sense beyond the discrete values. To clarify, for the given dataset $\{q^i, s_t^i,y^i\}^N$, people use the mean-squared error (MSE) loss to optimize value-based process verifier as follows:
\begin{equation}\label{reg:scalar}
    \min_\theta\sum_{i=1}^N(f_\theta(q, s_t)-y)^2
\end{equation}
where $f_\theta$ is the function mapping from state space $\{q\}\bigcup s_t\in\mathcal{S}$ to the contiguous space $y\in[0,1]$, $y$ is the estimated value from the current state to the outcome that ranges from 0 to 1. We name Eqn.\ref{reg:scalar} as scalar regression.

Similar to the \textbf{distance prior} injected via the mean-square error objective function, one can represent the contiguous value as the expectation of some categorical distribution, which is defined by the Dirac delta function and the category quantity. We name it as the \textbf{structural prior}. To clarify, we define the categorical distribution $\mathcal{Z}$ with $m$ locations as follows:
\begin{equation}
    \mathcal{Z}=\left\{\sum_{i=1}^{m}p_i\delta_{z_i}:p_i\geq0;\sum_{i=1}^{m}p_i=1\right\},
\end{equation}
where $z_i$ is the $i$-th location of the categorical distribution,  $p_i$ is the probability of selecting location $z_i$, and $\delta_{z_i}$ is the Dirac delta function at location $z_i$. The \textbf{structural prior} thus allow us to represent the $V^\pi(s)$ as the expectation of the categorical distribution as follows:
\begin{equation}\label{value_cate}
    V^\pi(s)=\mathbb{E}_p[\delta_i]=\sum_{i=1}^{m}p_i(s)\delta_i.
\end{equation}

Given $f_\theta$ mapping from state space $\{q\}\bigcup s_t\in\mathcal{S}$ to the categorical distribution $\mathcal{Z}\in\mathcal{R}^{m}$, we can also optimize value-based process verifier using the maximum likelihood estimator under MSE loss as follows:
\begin{equation}\label{reg:categorical}
    \mathcal{L}_\theta^{MSE}=\min_\theta\sum_{i=1}^N(f_\theta(q,s_t)\cdot\Delta_z^T-y)^2,
\end{equation}
where $\Delta_z=[\delta_{z_1},\delta_{z_2},...,\delta_{z_{m}}]$ is the weighting vector to calculate the expectation of categorical distribution $\mathcal{Z}$. We name Eqn.\ref{reg:categorical} as expectation regression.

\begin{algorithm}[tb]
   \caption{Regression Implementation}\label{alg:reg}
\begin{algorithmic}[1]\label{alg:reg}
   \STATE {\bfseries Input:} questions $Q$, answers $A$, states $S$, policy $\pi$, verifier $f_\theta$, Monte Carlo sampling size $k$
   \FOR{$\{q,a,s\}$ {\bfseries in} $\{Q,A,S\}$}
   \STATE Sample $k$ rollouts given $\pi(\cdot|q,s)$ and get $\hat{A}=\{\hat{a}_i\}_k$
   \STATE Estimate $\hat{V}^\pi(q,s)=\frac{\#\rm correct}{|\hat{A}|}$ by comparing ($\hat{a}_i$, $a$)
   \STATE Predict state value $\tilde{V}_\theta^\pi(q,s)$ using verifier $f_\theta$
   \IF{Scalar regression, Eqn.\ref{reg:scalar}:}
    \STATE $\tilde{V}_\theta^\pi(q,s)\leftarrow f_\theta(\cdot|q,s)$
   \ELSIF{Expectation regression, Eqn.\ref{reg:categorical}:}
    \STATE $\tilde{V}_\theta^\pi(q,s)\leftarrow f_\theta(\cdot|q,s)\cdot\Delta_z^T$ 
    as $\Delta_z^T$ defined in Eqn.\ref{reg:categorical}.
    \ENDIF
   \ENDFOR
   \STATE Optimize $f_\theta$ through MSE loss:
   \begin{equation}
       \min_\theta \sum_{(q,s)\in (Q,S)}(\tilde{V}_\theta^\pi(q,s)-\hat{V}^\pi(q,s))^2 \nonumber
   \end{equation}
\end{algorithmic}
\end{algorithm}

The na\"ive implementation while optimizing value-based process verifier using Eqn.\ref{reg:scalar} and Eqn.\ref{reg:categorical} is shown in Algorithm \ref{alg:reg}. Both methods calculate the estimated state value $\hat{V}^\pi(s)$ under Monte Carlo sampling, then perform MSE loss between the predicted value $f_\theta(\cdot,s)$ and estimated value $\hat{V}^\pi(s)$. Through sharing the same objective function and the identical distance prior, the expectation regression additionally follows the \textbf{structural prior}, which allows us to optimize the value-based process verifier from the distribution's perspective.

\subsection{Structural Prior Injection via Categorical Distribution Modeling}
By treating the k-times Monte Carlo sampling as a singular sample of the Monte Carlo estimation, the estimation distribution then follows the Binomial distribution:
\begin{equation}\label{mtd:mc_distribution}
    \hat{V}^\pi(s)\sim Bin(k,p),
\end{equation}
where $p$ is the ground-truth state value, $k$ is the sampling quantity. The scalar regression problem can thus be interpreted from the distribution perspective: given current state $s_t$, we sample once from the Binomial distribution $Bin(k,p)$, annotated as $V^\pi(s)$. Our goal is to estimate the expectation of the Binomial distribution $\mathbb{E}[V^\pi(s)]$ where $V^\pi(s)\sim Bin(k,p)$.  We name the value-based process verifier as expectation-based model.

We perform two different methods to optimize the value-based process verifier. The first method is through the mean-square error objective function, as we aim to recover the expectation of the Binomial distribution. We model the categorical distribution as having the identical Dirac delta function and category quantity compared with the Binomial distribution:
\begin{equation}
    \delta_{z_i}=z_i,z_i=\frac{i-1}{k},i\in[1,2,...,k+1],
\end{equation}
where $k$ is the sampling quantity, which is equal to the category quantity. We then optimize the expectation of the categorical distribution using Eqn.\ref{reg:categorical}. Compared with the scalar regression method, we inject the definition of the categorical distribution as the structural prior. 

The second method is through cross-entropy, or the Histogram Loss. Following the categorical distribution definition above, we aim to recover the ground-truth target distribution, i.e. the Binomial distribution $Bin(k,p)$. Ideally, if there's a chance to recover the ground-truth distribution given limited Monte Carlo sampling results, the expectation of the categorical distribution, which is the predicted state value, can be accurate and match the ground truth $p$. As it's difficult since we only have one sampled result from the Binomial distribution, the goal of reducing sampling error is to find the optimal posterior distribution class. To clarify, given limited sampling results, our goal is to find the optimal posterior distribution class that is conditioned on the sampling results, which has the minimum distance to the ground-truth Binomial distribution where the sampled results come from. Compared with the scalar regression method, the structural prior of the second method is two-fold. We first define the Dirac delta function and category quantity that is identical to the ground truth target distribution, then define the exact posterior distribution class that can be close to the ground truth target distribution. 

In the following section, we introduce the method to find the suitable posterior distribution class.

\subsection{Posterior Distribution Selection via Statistics-based Distance}
As the estimated state value $\hat{V}^\pi(s)$ follows the Binomial distribution as described in Eqn.\ref{mtd:mc_distribution}, the ground-truth target distribution of categorical distribution is the Binomial distribution with ground-truth success rate $p$ and number of Monte Carlo rollouts $k$. However, it is difficult to estimate the distribution with limited sampling instances in an unbiased way, especially when we sample only once from the categorical distribution. Given limited sampling results, we propose the metric to calculate the distance between posterior distribution and ground-truth target distribution at the statistical level as follows:
\begin{definition}
    \textit{Statistical Distribution Distance.} Given $p$ as the posterior distribution, $q$ as the target distribution, $d$ as one of the pre-defined distance metrics like Kullback-Leibler(KL) divergence, Wasserstein distance, or else. The distance between distributions at the statistical level is defined via statistics-based expectation as follows:
    \begin{equation}
        \mathcal{DT}(p,q)=\mathbb{E}_{x\sim q}[d(p(\cdot|x),q)].
    \end{equation}
\end{definition}
The KL divergence is useful to measure the distance between distributions. However, as the categories in the categorical distribution have sequence characteristics(i.e. have different weights), it's not suitable to measure the distance between categorical distributions by KL divergence. We use the Wasserstein distance instead. It's reasonable that the smaller the Statistical Distribution Distance is, the more accurately we estimate the target distribution, which will help improve the performance of the value-based process verifier.

After all, the value-based process verifier will be optimized by the Histogram Loss to provide distribution shaping supervise signal as follows:
\begin{equation}\label{mtd:hl}
    \mathcal{L}_\theta^{HL}=\min_\theta-\sum_{i=1}^N\sum_{j=1}^{|\mathcal{Z}|}f_\theta(z_j|q,s_t)\log p(z_j|q,s_t)
\end{equation}
where $j$ is the bin index of the categorical distribution $\mathcal{Z}$, $p(z_j)$ is the probability of the $j$-th bin of the posterior distribution $p$.
\section{Experiments}
\subsection{Experimental Settings}
\paragraph{Datasets and metrics.} Following previous research \citep{DBLP:conf/acl/WangLSXDLCWS24,DBLP:conf/iclr/LightmanKBEBLLS24}, we evaluate value-based process verifiers based on their verification ability through best-of-$N$ sampling. Specifically, given a problem $p$, we sample $N$ candidate solutions from a generator. Then, the candidates are re-ranked according to the score generated by the given verifier. The candidate with the highest score will be treated as the best candidate and selected as the final solution. Finally, we compare the consistency of the final solution and ground-truth answer to determine whether the solution is correct. The statistical success rate will be reported. Following \citet{DBLP:journals/corr/abs-2408-03314}, we also include the beam-search metric to evaluate the performance of value-based process verifiers at test time. Specifically, the number of beams $N$ and beam size $M$ will be set. Given a problem $p$, the verifier is required to score every step during the generation process. For each iteration, $N$ beams will generate $M$ next-step candidates individually, the verifier will select the top $N$ candidates from the $N*M$ beams as the beams for the next iteration, and then continue the iteration until finished. Like Best-of-N, we compare the consistency of the final solution and ground-truth answer. The statistical success rate will be reported.

We conduct our experiments on the challenging MATH dataset\citet{DBLP:conf/nips/HendrycksBKABTS21} for verification. We use the test split following \citet{DBLP:conf/iclr/LightmanKBEBLLS24} as our test set, which consists of 500 randomly selected questions from MATH. As described by \citet{DBLP:conf/acl/WangLSXDLCWS24}, the subset evaluation produces similar results to the full-set evaluation. We uses MetaMATH\citep{DBLP:conf/iclr/YuJSYLZKLWL24} as the fine-tuning dataset as used in \citet{DBLP:conf/acl/WangLSXDLCWS24}.

\paragraph{Baselines and implementation details.}
The generator in our experiments is based on LLemma-7b. We train it on MetaMATH for 3 epochs to get the generator. Based on the train split of MATH dataset, we construct the training dataset of the process verifier. To clarify, We use the generator to sample 15 solutions per problem. Following previous works\citep{DBLP:conf/iclr/LightmanKBEBLLS24, DBLP:conf/acl/WangLSXDLCWS24}, we split each solution into steps by the pre-defined rule-based strategies (e.g. newline as the delimiters). For each step, we combine it with its previous steps to form an incomplete solution, then sample 8 rollouts to perform Monte Carlo estimation and annotate the state value. In general, we sample $15*8=120$ samples for each problem and the training dataset has 180k samples in total.

We use the Qwen2.5-Math-7B-Instruct\citep{yang2024qwen2} and deepseek-math-7b-instruct\citep{DBLP:journals/corr/abs-2402-03300} as the value-based process verifier base model. For the scalar-regression methods, similar to \citet{DBLP:conf/acl/WangLSXDLCWS24}, we train the verifier in the language modeling way, adding special tokens to the model's vocabulary and use the probability of generating the positive token as the predicted state value. For the expectation-based methods, we add a linear layer then use the softmax function to map the categorical output into a probability distribution. We compare the following methods:
\begin{itemize}
    \item Scalar Regression (outcome). Scalar-regression model that is trained on the final state value.
    \item Scalar Regression. Scalar-regression model that is trained on every intermediate state value, including the final state value.
    \item Expectation Regression (MSE). Expectation-based model that trained with mean-square error objective function on every intermediate state value, including the final state value.
    \item Expectation Classification (HL). Expectation-based model that trained with Histogram Loss objective function on a pre-defined distribution on every intermediate state value, including the final state value.
\end{itemize}

For the Expectation Classification method, we use the truncated Gaussian distribution where $\mu$ is the sampled state value, $\sigma^2$ is $\frac{\mu(1-\mu)}{k}$, as the Gaussian distribution can be used as an approximation of the binomial distribution, which can result in a more precise posterior distribution and thus reduce distribution mismatch. A further analysis of distribution selection can be found in section\ref{analysis_sampling}.

\subsection{Results}
\begin{table*}[h!]
\small
\centering
\caption{The performance comparison of PRMs. The numbers represent the percentage of problems solved using verifiers. For all BoN experiments, we use the final state value to re-rank solution candidates following the definition of state value in LLM reasoning scenario.}
\label{tab:experiment-results}
\begin{tabular}{l|ccccc|cc}
\toprule
\multirow{2}{*}{Method}                           & \multicolumn{5}{c|}{Best-of-N}    & \multicolumn{2}{c}{Beam Search}   \\
% \multirow{2}{*}{Method}                           & \multicolumn{5}{c|}{Best-of-N}    & Beam Search   \\
                                & 8     & 16    & 32    & 64       & 128         & 4 & 8             \\ \midrule
Oracle                          & 29.8  & 35.8  & 40.4     & 45.7     & 49.6   & - & -                   \\
\rowcolor{gray!12}
\multicolumn{8}{c}{\textbf{Qwen2.5-Math-7B-Instruct}} \\
Scalar Regression (outcome)     & 24.4  & 28.2  & 29.4  & 30.0  & 31.2    & 44.2 & 45.8              \\
Scalar Regression (process)     & 24.4  & 28.2  & 29.0  & 29.8  & 30.6     & 51.8 & 56.0              \\
Expectation Regression (MSE)    & 24.4  & 27.8  & 29.0  & \textbf{30.6}  & \textbf{31.8}    & \textbf{52.0} & \textbf{56.4}               \\
Expectation Classification (HL) & \textbf{24.8}  & \textbf{28.4}  & \textbf{29.6}  & 30.4  & \textbf{31.8}   & 51.8 & 56.0                 \\ \midrule
\rowcolor{gray!12}
\multicolumn{8}{c}{\textbf{Deepseek-math-7b-instruct}} \\
Scalar Regression (outcome)     & 11.8  & 12.6  & 12.0  & 11.6   & 13.2    & 17.0 & 14.0              \\
Scalar Regression (process)     & 21.0  & 23.2  & 23.0  & 23.6  & 24.4     & 43.6 & 46.2              \\
Expectation Regression (MSE)    & 22.0  & \textbf{23.4}  & 25.0  & \textbf{25.8}  & 25.8     & \textbf{44.0} & 46.0                \\
Expectation Classification (HL) & \textbf{22.4}  & \textbf{23.4} & \textbf{25.4}  & 25.4  & \textbf{26.4}     & 42.6 & \textbf{47.0}              \\ \midrule
\end{tabular}
\end{table*}

For the BoN experiments, we report the Best-of-N results and $N\in\{8,16,32,64,128\}$. For the beam search experiments, we report the result of both the number of beams and beam size are the same value, $M=N\in\{4,8\}$. The results are shown in Table \ref{tab:experiment-results}. Under the categorical distribution definition, the expectation-based models that directly optimized the categorical distribution by its expectation, or incorporating the pre-defined Gaussian distribution then directly optimized the shape of the categorical distribution using Histogram loss, perform consistently better than the scalar regression baseline, showing that the structural prior injection method can effectively improve the performance of process verifier trained via different optimization objective functions. The results also indicate that under reasonable structural prior, the expectation-based method can be an elegant replacement for the scalar-regression verifier at little-to-no cost.

We also see a superior performance of the outcome-supervised verifier compared with its process-supervised version on the Best-of-N task, when using Qwen2.5-Math-7B-Instruct as the base model of the verifier. One possible reason is that the value annotations are accurate for the final states. For the intermediate states, the labels contain error and noise due to Monte Carlo estimation and thus result in inferior performance. We provide more information about the Monte Carlo error analysis in Appendix \ref{app:noise}. However, though the outcome-supervised verifier has satisfied performance for the Best-of-N task, it doesn't perform well in Beam Search, which shows the importance of process supervision and the broad advantages of our method.
\section{Analysis and Discussion}
In this section, we will discuss the influence of the Statistics-based Distance metric and the structural prior. Though related to each other, we discuss the distance metric and structural prior in different sections for the sake of clarity. The former is based on the cross-entropy loss that can be directly supervised by specific types of categorical distribution definitions. The latter is based on the mean-square error objective function where the training objective is only supervised on the expectation of the categorical distribution, which allows us to define the distribution more freely.
\subsection{Posterior Distribution Selection.}\label{analysis_sampling}
\begin{figure*}[h!]
    \centering
    \begin{subfigure}[t]{0.3\textwidth}
        \centering
        \includegraphics[width=\textwidth]{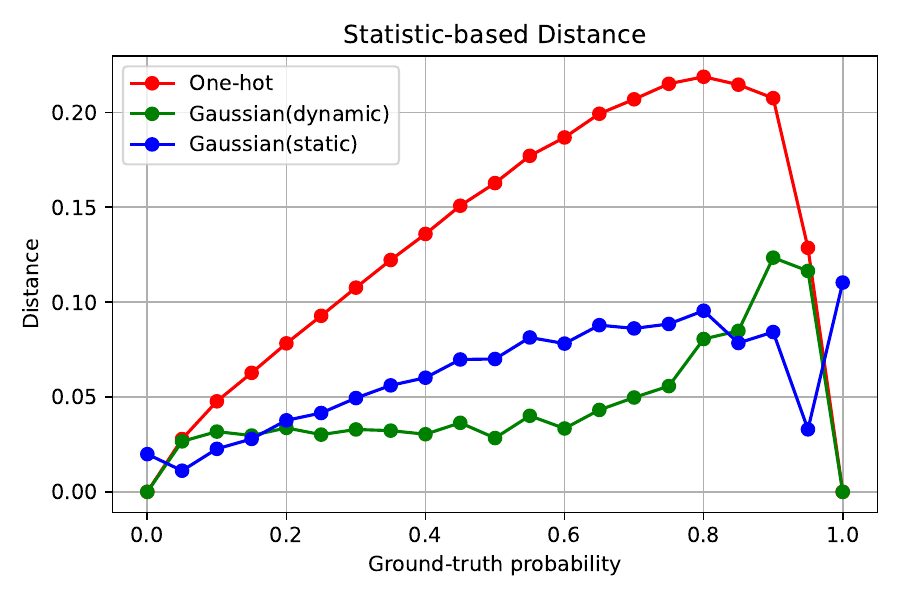}
        \caption{The statistics-based distances between the ground-truth distribution and the pre-defined posterior distribution.}
        \label{fig:distance}
    \end{subfigure}
    \hfill
    \begin{subfigure}[t]{0.3\textwidth}
        \centering
        \includegraphics[width=\textwidth]{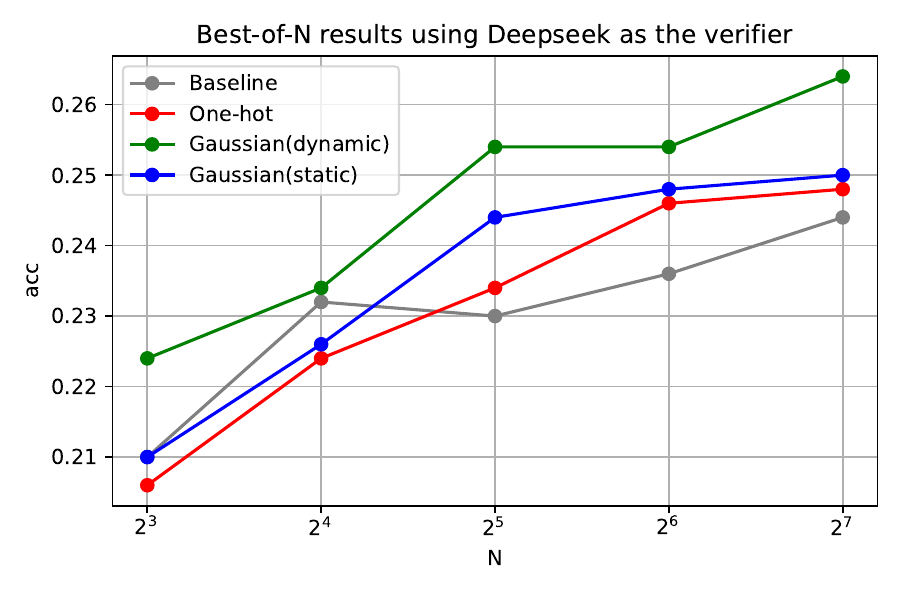}
        \caption{The Best-of-N results when selection different posterior distributions and using Deepseek-math-7b-instruct as the verifier.}
        \label{fig:bon}
    \end{subfigure}
    \hfill
    \begin{subfigure}[t]{0.3\textwidth}
        \centering
        \includegraphics[width=\textwidth]{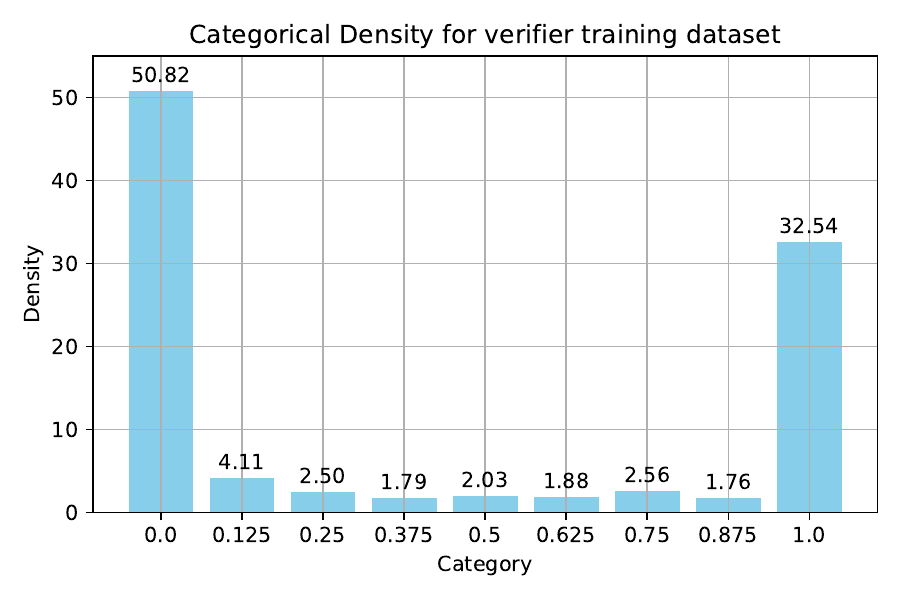}
        \caption{The actual categorical distribution of our training dataset.}
        \label{fig:distribution}
    \end{subfigure}
    \caption{The ablation study about the posterior distribution selection. We compare different posterior distributions, measuring them based on the statistics-based distance, and the Best-of-N performance in MATH500.}
\end{figure*}

In this section, we analyze the difference in distribution selection and their corresponding performance. To clarify, we use the following distributions:
\begin{itemize}
    \item One-hot distribution. A one-hot distribution that the category which sampled state value points to is set to be 1, other categories are set to 0.
    \item Guassian distribution(dynamic). Truncated Gaussian distribution with dynamic variance. $\mu$ is the sampled state value, $\sigma^2$ is $\frac{\mu(1-\mu)}{k}$.
    \item Gaussian distribution(static). Truncated Gaussian distribution with fixed variance as described in \citet{DBLP:conf/icml/FarebrotherOVTC24}. We set three standard deviations to be two bin widths in our experiments.
\end{itemize}
We report the statistics-based distance for different posterior distributions in Figure \ref{fig:distance}. Specifically, we calculate the statistics-based distance varying on different ground-truth probability $p$ from 0 to 1. We use the Wasserstein distance to measure the distance between categorical distributions. The distance between the $i$-th category and $j$-th category is $\frac{|i-j|}{k}$ following the definition of our categorical distribution. As shown in Figure \ref{fig:distance}, compared with Gaussian distributions, one-hot distribution has a much greater distance varying on the ground-truth probability $p$. For the Gaussian(dynamic) distribution, the statistics-based distance is smaller than Gaussian(static) distribution in most cases varying on the ground-truth probability $p$.

We report the performance of value-based process verifiers incorporating each posterior distribution based on Deepseek-math-7b-instruct in Figure \ref{fig:bon}. Compared with the baseline scalar regression method, incorporating different distributions can effectively improve the performance. As shown in the figure, the improvement is relatively more impressive when incorporating Gaussian distributions, especially the Gaussian(dynamic) distribution compared with the one-hot distribution. The results show that the statistics-based distance can be a good metric for designing the posterior distribution and avoiding distribution mismatch. 

Finally, we report the global estimated value distribution in Figure \ref{fig:distribution}, as a proxy to the ground-truth probability distribution which is actually intractable. As shown in the figure, the estimated value distribution is not an uniform distribution but clustered at marginal values, which may indicate that the ground-truth probability distribution is not flat but relatively extreme. Combining Figure \ref{fig:distance} and Figure \ref{fig:distribution}, the reason why the performances are slightly different between verifiers incorporating different distributions may be because of their similar statistics-based distance for $p$ closed to marginal values.

\subsection{Prior Categorical Distribution Selection.}
\begin{figure}[h!]
    \centering
    \begin{subfigure}[t]{0.23\textwidth}
        \centering
        \includegraphics[width=\textwidth]{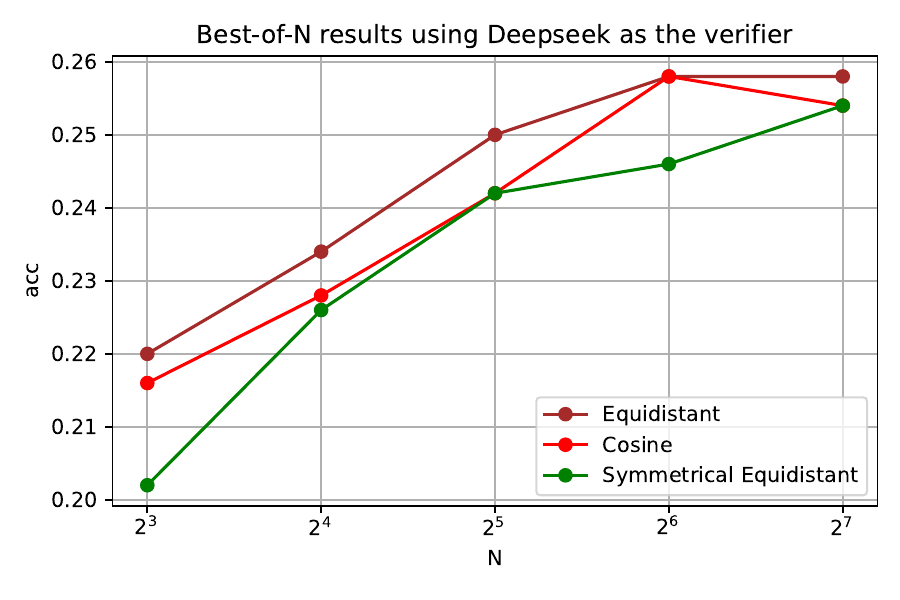}
    \end{subfigure}
    \hfill
    \begin{subfigure}[t]{0.23\textwidth}
        \centering
        \includegraphics[width=\textwidth]{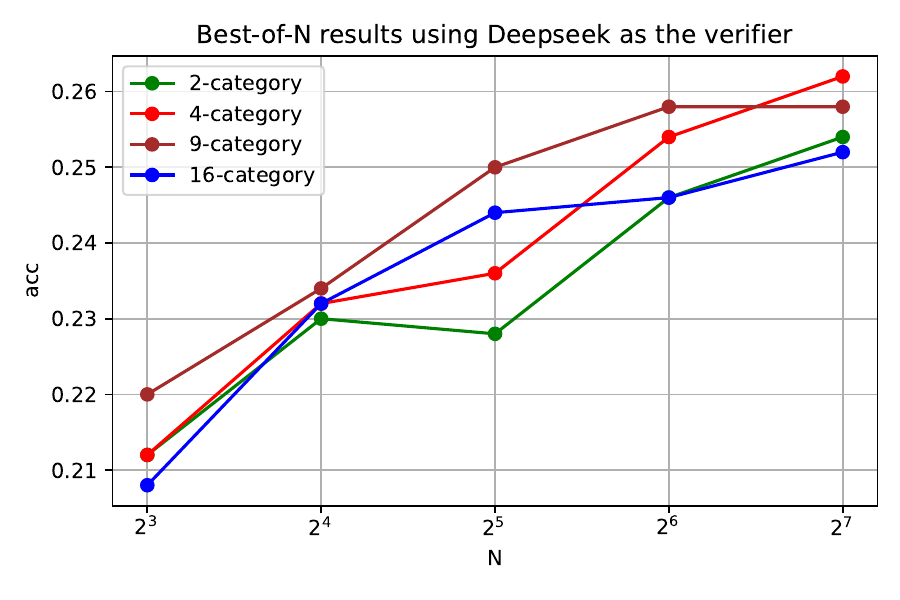}
    \end{subfigure}
    \caption{The ablation study about the categorical distribution selection. We compare different categorical distributions varying on the Dirac delta function definition and the category quantity. We report the Best-of-N performance in MATH500.}
    \label{fig:mse}
\end{figure}

In this section, we analyze the difference in the categorical definition and their corresponding performance, thus discussing the influence of prior structural information to be injected into the verifier. To clarify, we compare the Dirac delta function definitions and the category quantities to construct the categorical distribution, then optimize the distribution under the mean-square error objective function. Following the definition of $\delta_{z_i}$ in sec \ref{sec:regression}, let $\delta_{z_i}$ be the Dirac delta function at location $z_i$, where $i\in[1,2,...,k+1]$.  We compare the following Dirac delta functions:
\begin{itemize}
    \item Equidistant. $\delta_{z_i}=\frac{i-1}{k}$.
    \item Cosine. $\delta_{z_i}=1-cos(\frac{i-1}{k}\pi)$.
    \item Symmetrical Equidistant. $\delta_{z_i}=-1+\frac{2(i-1)}{k}$
\end{itemize}
We fixed the category quantity to be 9, then the categorical distribution using the equidistant Dirac delta function can be treated as formulating a Binomial distribution with a sample size of $8$. The other Dirac delta functions can be treated as the varieties of the equidistant Dirac delta function, though we preserve the monotonicity, upper and lower bound of the equidistant Dirac delta function. Apart from the different Dirac delta function definitions, we also explore the difference in category quantities. We test the category quantity to be {2,4,9,16} and for all categorical distributions that varying on category quantities, we use the equidistant Dirac delta function. For both experiments, we report the Best-of-N result in Figure \ref{fig:mse}.

Our results show that the structural prior can effectively influence the performance of value-based process verifiers. Though preserving similar features of the equidistant Dirac delta function, the other Dirac delta functions are more difficult to provide meaningful prior information and thus the optimization process will be more difficult. Similar results can be found in the experiment of varieties of category quantity. It shows that setting the category quantity to match the Binomial distribution is useful for better performance.

\section{Limitation and Future work}
We believe that the structural prior injection method, as well as the categorical distribution modeling method, provides a meaningful aspect to improve the performance of the value-based process verifier and analyze the Monte Carlo estimation error. However, we are certain that research about the structural prior still has a long way to go. For instance, when optimizing the categorical distribution via Histogram Loss, as we use the statistics-based distance to measure the distance between the posterior distribution and ground-truth distribution, the distribution modeling can only be performed in a prior and non-differentiable way due to the non-differentiable nature of Wasserstein distance. One future direction is to soften the distance metric and transform the distribution modeling approach into a tractable and dynamic way. 

The other thing is that we use the Histogram Loss to optimize the categorical distribution representation, which is not a perfect objective function due to the absence of distance prior. To clarify, as the categorical distribution is defined under the Dirac delta function of unsteady values, the categories then follow a natural order since our goal is the expectation of the categorical distribution. When optimizing the verifier via Histogram Loss, the distance between different categories will not be taken into account. Like other injected prior information, the data prior will not change the optimal solution which is the estimated posterior distribution, but it should result in a better performance for reasons like faster convergence or more accurate objective.

Finally, while our work discusses the mean-square error and cross-entropy loss from the unified perspective, we find it hard to combine the two different objective functions together. When training verifiers using both the mean-square error and cross-entropy loss by their linear combinations, we find there's a performance drop compared with training on each objective function individually. The possible reason may be the different gradient directions during the optimization process or the different assumptions from the perspective of maximum likelihood optimization. Further work can perform more analysis of the potential and scalability of using both objective functions together.

\section{Conclusion}
In this paper, we introduce the \textbf{structural prior}, transforming the scalar value into the expectation of a pre-defined categorical distribution, which allows us to improve the value representation and address the Monte Carlo estimation error from the perspective of distribution estimation. Under suitable structural prior, we show that the error is derived from the mismatch between the estimated posterior distribution and the ground-truth distribution, which is intractable to estimate as we have only limited samples. We then provide the Statistics-based Distance as the metric to measure the distance between ground truth distribution and posterior distribution. Under the vanilla mean-square error objective function and Histogram Loss objective function, our experiments show that the transformation can yield consistent improvements in performance on different tasks in the LLM reasoning scenario, showing that the structural prior can be of great benefit to the optimization process of the value-based process verifier. Finally, we compare the effectiveness varyin on structural priors. By performing experiments on each objective function, we show that prior selection can greatly influence the performance of value-based process verifiers.

\section*{Impact Statements}
This paper presents a method whose goal is to advance the large language model reasoning. This endeavor, while technically challenging, carries significant implications for the ethical use and societal impact of artificial intelligence. The goal of model's reasoning ability is to provide a more reasonable and useful tool for the human's benefit. It enhances AI’s utility in various sectors to make reliable decisions that are in line with organizational goals and ethical standards.

\bibliography{ref}

@inproceedings{DBLP:conf/acl/WangLSXDLCWS24,
  author       = {Peiyi Wang and
                  Lei Li and
                  Zhihong Shao and
                  Runxin Xu and
                  Damai Dai and
                  Yifei Li and
                  Deli Chen and
                  Yu Wu and
                  Zhifang Sui},
  editor       = {Lun{-}Wei Ku and
                  Andre Martins and
                  Vivek Srikumar},
  title        = {Math-Shepherd: Verify and Reinforce LLMs Step-by-step without Human
                  Annotations},
  booktitle    = {Proceedings of the 62nd Annual Meeting of the Association for Computational
                  Linguistics (Volume 1: Long Papers), {ACL} 2024, Bangkok, Thailand,
                  August 11-16, 2024},
  pages        = {9426--9439},
  publisher    = {Association for Computational Linguistics},
  year         = {2024},
  url          = {https://doi.org/10.18653/v1/2024.acl-long.510},
  doi          = {10.18653/V1/2024.ACL-LONG.510},
  bibsource    = {dblp computer science bibliography, https://dblp.org}
}

@inproceedings{DBLP:conf/iclr/LightmanKBEBLLS24,
  author       = {Hunter Lightman and
                  Vineet Kosaraju and
                  Yuri Burda and
                  Harrison Edwards and
                  Bowen Baker and
                  Teddy Lee and
                  Jan Leike and
                  John Schulman and
                  Ilya Sutskever and
                  Karl Cobbe},
  title        = {Let's Verify Step by Step},
  booktitle    = {The Twelfth International Conference on Learning Representations,
                  {ICLR} 2024, Vienna, Austria, May 7-11, 2024},
  publisher    = {OpenReview.net},
  year         = {2024},
  url          = {https://openreview.net/forum?id=v8L0pN6EOi},
  bibsource    = {dblp computer science bibliography, https://dblp.org}
}

@article{DBLP:journals/corr/abs-2406-06592,
  author       = {Liangchen Luo and
                  Yinxiao Liu and
                  Rosanne Liu and
                  Samrat Phatale and
                  Harsh Lara and
                  Yunxuan Li and
                  Lei Shu and
                  Yun Zhu and
                  Lei Meng and
                  Jiao Sun and
                  Abhinav Rastogi},
  title        = {Improve Mathematical Reasoning in Language Models by Automated Process
                  Supervision},
  journal      = {CoRR},
  volume       = {abs/2406.06592},
  year         = {2024},
  url          = {https://doi.org/10.48550/arXiv.2406.06592},
  doi          = {10.48550/ARXIV.2406.06592},
  eprinttype    = {arXiv},
  eprint       = {2406.06592},
  bibsource    = {dblp computer science bibliography, https://dblp.org}
}

@article{DBLP:journals/corr/abs-2408-03314,
  author       = {Charlie Snell and
                  Jaehoon Lee and
                  Kelvin Xu and
                  Aviral Kumar},
  title        = {Scaling {LLM} Test-Time Compute Optimally can be More Effective than
                  Scaling Model Parameters},
  journal      = {CoRR},
  volume       = {abs/2408.03314},
  year         = {2024},
  url          = {https://doi.org/10.48550/arXiv.2408.03314},
  doi          = {10.48550/ARXIV.2408.03314},
  eprinttype    = {arXiv},
  eprint       = {2408.03314},
  bibsource    = {dblp computer science bibliography, https://dblp.org}
}

@inproceedings{DBLP:conf/nips/HendrycksBKABTS21,
  author       = {Dan Hendrycks and
                  Collin Burns and
                  Saurav Kadavath and
                  Akul Arora and
                  Steven Basart and
                  Eric Tang and
                  Dawn Song and
                  Jacob Steinhardt},
  editor       = {Joaquin Vanschoren and
                  Sai{-}Kit Yeung},
  title        = {Measuring Mathematical Problem Solving With the {MATH} Dataset},
  booktitle    = {Proceedings of the Neural Information Processing Systems Track on
                  Datasets and Benchmarks 1, NeurIPS Datasets and Benchmarks 2021, December
                  2021, virtual},
  year         = {2021},
  url          = {https://datasets-benchmarks-proceedings.neurips.cc/paper/2021/hash/be83ab3ecd0db773eb2dc1b0a17836a1-Abstract-round2.html},
  bibsource    = {dblp computer science bibliography, https://dblp.org}
}

@inproceedings{DBLP:conf/iclr/YuJSYLZKLWL24,
  author       = {Longhui Yu and
                  Weisen Jiang and
                  Han Shi and
                  Jincheng Yu and
                  Zhengying Liu and
                  Yu Zhang and
                  James T. Kwok and
                  Zhenguo Li and
                  Adrian Weller and
                  Weiyang Liu},
  title        = {MetaMath: Bootstrap Your Own Mathematical Questions for Large Language
                  Models},
  booktitle    = {The Twelfth International Conference on Learning Representations,
                  {ICLR} 2024, Vienna, Austria, May 7-11, 2024},
  publisher    = {OpenReview.net},
  year         = {2024},
  url          = {https://openreview.net/forum?id=N8N0hgNDRt},
  bibsource    = {dblp computer science bibliography, https://dblp.org}
}

@inproceedings{DBLP:conf/icml/FarebrotherOVTC24,
  author       = {Jesse Farebrother and
                  Jordi Orbay and
                  Quan Vuong and
                  Adrien Ali Ta{\"{\i}}ga and
                  Yevgen Chebotar and
                  Ted Xiao and
                  Alex Irpan and
                  Sergey Levine and
                  Pablo Samuel Castro and
                  Aleksandra Faust and
                  Aviral Kumar and
                  Rishabh Agarwal},
  title        = {Stop Regressing: Training Value Functions via Classification for Scalable
                  Deep {RL}},
  booktitle    = {Forty-first International Conference on Machine Learning, {ICML} 2024,
                  Vienna, Austria, July 21-27, 2024},
  publisher    = {OpenReview.net},
  year         = {2024},
  url          = {https://openreview.net/forum?id=dVpFKfqF3R},
  bibsource    = {dblp computer science bibliography, https://dblp.org}
}

@inproceedings{DBLP:conf/nips/Wei0SBIXCLZ22,
  author       = {Jason Wei and
                  Xuezhi Wang and
                  Dale Schuurmans and
                  Maarten Bosma and
                  Brian Ichter and
                  Fei Xia and
                  Ed H. Chi and
                  Quoc V. Le and
                  Denny Zhou},
  editor       = {Sanmi Koyejo and
                  S. Mohamed and
                  A. Agarwal and
                  Danielle Belgrave and
                  K. Cho and
                  A. Oh},
  title        = {Chain-of-Thought Prompting Elicits Reasoning in Large Language Models},
  booktitle    = {Advances in Neural Information Processing Systems 35: Annual Conference
                  on Neural Information Processing Systems 2022, NeurIPS 2022, New Orleans,
                  LA, USA, November 28 - December 9, 2022},
  year         = {2022},
  url          = {http://papers.nips.cc/paper\_files/paper/2022/hash/9d5609613524ecf4f15af0f7b31abca4-Abstract-Conference.html},
  bibsource    = {dblp computer science bibliography, https://dblp.org}
}

@article{DBLP:journals/corr/abs-2211-09066,
  author       = {Hattie Zhou and
                  Azade Nova and
                  Hugo Larochelle and
                  Aaron C. Courville and
                  Behnam Neyshabur and
                  Hanie Sedghi},
  title        = {Teaching Algorithmic Reasoning via In-context Learning},
  journal      = {CoRR},
  volume       = {abs/2211.09066},
  year         = {2022},
  url          = {https://doi.org/10.48550/arXiv.2211.09066},
  doi          = {10.48550/ARXIV.2211.09066},
  eprinttype    = {arXiv},
  eprint       = {2211.09066},
  bibsource    = {dblp computer science bibliography, https://dblp.org}
}

@inproceedings{qin-etal-2024-context,
    title = "In-Context Learning with Iterative Demonstration Selection",
    author = "Qin, Chengwei  and
      Zhang, Aston  and
      Chen, Chen  and
      Dagar, Anirudh  and
      Ye, Wenming",
    editor = "Al-Onaizan, Yaser  and
      Bansal, Mohit  and
      Chen, Yun-Nung",
    booktitle = "Findings of the Association for Computational Linguistics: EMNLP 2024",
    month = nov,
    year = "2024",
    address = "Miami, Florida, USA",
    publisher = "Association for Computational Linguistics",
    url = "https://aclanthology.org/2024.findings-emnlp.438/",
    doi = "10.18653/v1/2024.findings-emnlp.438",
    pages = "7441--7455"
}

@inproceedings{DBLP:conf/nips/YaoYZS00N23,
  author       = {Shunyu Yao and
                  Dian Yu and
                  Jeffrey Zhao and
                  Izhak Shafran and
                  Tom Griffiths and
                  Yuan Cao and
                  Karthik Narasimhan},
  editor       = {Alice Oh and
                  Tristan Naumann and
                  Amir Globerson and
                  Kate Saenko and
                  Moritz Hardt and
                  Sergey Levine},
  title        = {Tree of Thoughts: Deliberate Problem Solving with Large Language Models},
  booktitle    = {Advances in Neural Information Processing Systems 36: Annual Conference
                  on Neural Information Processing Systems 2023, NeurIPS 2023, New Orleans,
                  LA, USA, December 10 - 16, 2023},
  year         = {2023},
  url          = {http://papers.nips.cc/paper\_files/paper/2023/hash/271db9922b8d1f4dd7aaef84ed5ac703-Abstract-Conference.html},
  bibsource    = {dblp computer science bibliography, https://dblp.org}
}

@inproceedings{DBLP:conf/emnlp/LiuG0HZQZ23,
  author       = {Tengxiao Liu and
                  Qipeng Guo and
                  Yuqing Yang and
                  Xiangkun Hu and
                  Yue Zhang and
                  Xipeng Qiu and
                  Zheng Zhang},
  editor       = {Houda Bouamor and
                  Juan Pino and
                  Kalika Bali},
  title        = {Plan, Verify and Switch: Integrated Reasoning with Diverse X-of-Thoughts},
  booktitle    = {Proceedings of the 2023 Conference on Empirical Methods in Natural
                  Language Processing, {EMNLP} 2023, Singapore, December 6-10, 2023},
  pages        = {2807--2822},
  publisher    = {Association for Computational Linguistics},
  year         = {2023},
  url          = {https://doi.org/10.18653/v1/2023.emnlp-main.169},
  doi          = {10.18653/V1/2023.EMNLP-MAIN.169},
  bibsource    = {dblp computer science bibliography, https://dblp.org}
}

@inproceedings{DBLP:conf/iclr/0002WSLCNCZ23,
  author       = {Xuezhi Wang and
                  Jason Wei and
                  Dale Schuurmans and
                  Quoc V. Le and
                  Ed H. Chi and
                  Sharan Narang and
                  Aakanksha Chowdhery and
                  Denny Zhou},
  title        = {Self-Consistency Improves Chain of Thought Reasoning in Language Models},
  booktitle    = {The Eleventh International Conference on Learning Representations,
                  {ICLR} 2023, Kigali, Rwanda, May 1-5, 2023},
  publisher    = {OpenReview.net},
  year         = {2023},
  url          = {https://openreview.net/forum?id=1PL1NIMMrw},
  bibsource    = {dblp computer science bibliography, https://dblp.org}
}

@inproceedings{DBLP:conf/emnlp/ZhaoXKHX23,
  author       = {James Xu Zhao and
                  Yuxi Xie and
                  Kenji Kawaguchi and
                  Junxian He and
                  Michael Qizhe Xie},
  editor       = {Houda Bouamor and
                  Juan Pino and
                  Kalika Bali},
  title        = {Automatic Model Selection with Large Language Models for Reasoning},
  booktitle    = {Findings of the Association for Computational Linguistics: {EMNLP}
                  2023, Singapore, December 6-10, 2023},
  pages        = {758--783},
  publisher    = {Association for Computational Linguistics},
  year         = {2023},
  url          = {https://doi.org/10.18653/v1/2023.findings-emnlp.55},
  doi          = {10.18653/V1/2023.FINDINGS-EMNLP.55},
  bibsource    = {dblp computer science bibliography, https://dblp.org}
}

@article{DBLP:journals/corr/abs-2409-12183,
  author       = {Zayne Sprague and
                  Fangcong Yin and
                  Juan Diego Rodriguez and
                  Dongwei Jiang and
                  Manya Wadhwa and
                  Prasann Singhal and
                  Xinyu Zhao and
                  Xi Ye and
                  Kyle Mahowald and
                  Greg Durrett},
  title        = {To CoT or not to CoT? Chain-of-thought helps mainly on math and symbolic
                  reasoning},
  journal      = {CoRR},
  volume       = {abs/2409.12183},
  year         = {2024},
  url          = {https://doi.org/10.48550/arXiv.2409.12183},
  doi          = {10.48550/ARXIV.2409.12183},
  eprinttype    = {arXiv},
  eprint       = {2409.12183},
  bibsource    = {dblp computer science bibliography, https://dblp.org}
}

@misc{deepseekai2025deepseekr1incentivizingreasoningcapability,
      title={DeepSeek-R1: Incentivizing Reasoning Capability in LLMs via Reinforcement Learning}, 
      author={DeepSeek-AI and Daya Guo and Dejian Yang and Haowei Zhang and Junxiao Song and Ruoyu Zhang and Runxin Xu and Qihao Zhu and Shirong Ma and Peiyi Wang and Xiao Bi and Xiaokang Zhang and Xingkai Yu and Yu Wu and Z. F. Wu and Zhibin Gou and Zhihong Shao and Zhuoshu Li and Ziyi Gao and Aixin Liu and Bing Xue and Bingxuan Wang and Bochao Wu and Bei Feng and Chengda Lu and Chenggang Zhao and Chengqi Deng and Chenyu Zhang and Chong Ruan and Damai Dai and Deli Chen and Dongjie Ji and Erhang Li and Fangyun Lin and Fucong Dai and Fuli Luo and Guangbo Hao and Guanting Chen and Guowei Li and H. Zhang and Han Bao and Hanwei Xu and Haocheng Wang and Honghui Ding and Huajian Xin and Huazuo Gao and Hui Qu and Hui Li and Jianzhong Guo and Jiashi Li and Jiawei Wang and Jingchang Chen and Jingyang Yuan and Junjie Qiu and Junlong Li and J. L. Cai and Jiaqi Ni and Jian Liang and Jin Chen and Kai Dong and Kai Hu and Kaige Gao and Kang Guan and Kexin Huang and Kuai Yu and Lean Wang and Lecong Zhang and Liang Zhao and Litong Wang and Liyue Zhang and Lei Xu and Leyi Xia and Mingchuan Zhang and Minghua Zhang and Minghui Tang and Meng Li and Miaojun Wang and Mingming Li and Ning Tian and Panpan Huang and Peng Zhang and Qiancheng Wang and Qinyu Chen and Qiushi Du and Ruiqi Ge and Ruisong Zhang and Ruizhe Pan and Runji Wang and R. J. Chen and R. L. Jin and Ruyi Chen and Shanghao Lu and Shangyan Zhou and Shanhuang Chen and Shengfeng Ye and Shiyu Wang and Shuiping Yu and Shunfeng Zhou and Shuting Pan and S. S. Li and Shuang Zhou and Shaoqing Wu and Shengfeng Ye and Tao Yun and Tian Pei and Tianyu Sun and T. Wang and Wangding Zeng and Wanjia Zhao and Wen Liu and Wenfeng Liang and Wenjun Gao and Wenqin Yu and Wentao Zhang and W. L. Xiao and Wei An and Xiaodong Liu and Xiaohan Wang and Xiaokang Chen and Xiaotao Nie and Xin Cheng and Xin Liu and Xin Xie and Xingchao Liu and Xinyu Yang and Xinyuan Li and Xuecheng Su and Xuheng Lin and X. Q. Li and Xiangyue Jin and Xiaojin Shen and Xiaosha Chen and Xiaowen Sun and Xiaoxiang Wang and Xinnan Song and Xinyi Zhou and Xianzu Wang and Xinxia Shan and Y. K. Li and Y. Q. Wang and Y. X. Wei and Yang Zhang and Yanhong Xu and Yao Li and Yao Zhao and Yaofeng Sun and Yaohui Wang and Yi Yu and Yichao Zhang and Yifan Shi and Yiliang Xiong and Ying He and Yishi Piao and Yisong Wang and Yixuan Tan and Yiyang Ma and Yiyuan Liu and Yongqiang Guo and Yuan Ou and Yuduan Wang and Yue Gong and Yuheng Zou and Yujia He and Yunfan Xiong and Yuxiang Luo and Yuxiang You and Yuxuan Liu and Yuyang Zhou and Y. X. Zhu and Yanhong Xu and Yanping Huang and Yaohui Li and Yi Zheng and Yuchen Zhu and Yunxian Ma and Ying Tang and Yukun Zha and Yuting Yan and Z. Z. Ren and Zehui Ren and Zhangli Sha and Zhe Fu and Zhean Xu and Zhenda Xie and Zhengyan Zhang and Zhewen Hao and Zhicheng Ma and Zhigang Yan and Zhiyu Wu and Zihui Gu and Zijia Zhu and Zijun Liu and Zilin Li and Ziwei Xie and Ziyang Song and Zizheng Pan and Zhen Huang and Zhipeng Xu and Zhongyu Zhang and Zhen Zhang},
      year={2025},
      eprint={2501.12948},
      archivePrefix={arXiv},
      primaryClass={cs.CL},
      url={https://arxiv.org/abs/2501.12948}, 
}

@inproceedings{DBLP:conf/nips/ZelikmanWMG22,
  author       = {Eric Zelikman and
                  Yuhuai Wu and
                  Jesse Mu and
                  Noah D. Goodman},
  editor       = {Sanmi Koyejo and
                  S. Mohamed and
                  A. Agarwal and
                  Danielle Belgrave and
                  K. Cho and
                  A. Oh},
  title        = {STaR: Bootstrapping Reasoning With Reasoning},
  booktitle    = {Advances in Neural Information Processing Systems 35: Annual Conference
                  on Neural Information Processing Systems 2022, NeurIPS 2022, New Orleans,
                  LA, USA, November 28 - December 9, 2022},
  year         = {2022},
  url          = {http://papers.nips.cc/paper\_files/paper/2022/hash/639a9a172c044fbb64175b5fad42e9a5-Abstract-Conference.html},
  bibsource    = {dblp computer science bibliography, https://dblp.org}
}

@article{DBLP:journals/corr/abs-2412-06559,
  author       = {Chujie Zheng and
                  Zhenru Zhang and
                  Beichen Zhang and
                  Runji Lin and
                  Keming Lu and
                  Bowen Yu and
                  Dayiheng Liu and
                  Jingren Zhou and
                  Junyang Lin},
  title        = {ProcessBench: Identifying Process Errors in Mathematical Reasoning},
  journal      = {CoRR},
  volume       = {abs/2412.06559},
  year         = {2024},
  url          = {https://doi.org/10.48550/arXiv.2412.06559},
  doi          = {10.48550/ARXIV.2412.06559},
  eprinttype    = {arXiv},
  eprint       = {2412.06559},
  bibsource    = {dblp computer science bibliography, https://dblp.org}
}

@inproceedings{DBLP:conf/icml/XiCHJZHDLGWGSFZ24,
  author       = {Zhiheng Xi and
                  Wenxiang Chen and
                  Boyang Hong and
                  Senjie Jin and
                  Rui Zheng and
                  Wei He and
                  Yiwen Ding and
                  Shichun Liu and
                  Xin Guo and
                  Junzhe Wang and
                  Honglin Guo and
                  Wei Shen and
                  Xiaoran Fan and
                  Yuhao Zhou and
                  Shihan Dou and
                  Xiao Wang and
                  Xinbo Zhang and
                  Peng Sun and
                  Tao Gui and
                  Qi Zhang and
                  Xuanjing Huang},
  title        = {Training Large Language Models for Reasoning through Reverse Curriculum
                  Reinforcement Learning},
  booktitle    = {Forty-first International Conference on Machine Learning, {ICML} 2024,
                  Vienna, Austria, July 21-27, 2024},
  publisher    = {OpenReview.net},
  year         = {2024},
  url          = {https://openreview.net/forum?id=t82Y3fmRtk},
  bibsource    = {dblp computer science bibliography, https://dblp.org}
}

@inproceedings{lu2024autopsv,
  title={AutoPSV: Automated Process-Supervised Verifier},
  author={Lu, Jianqiao and Dou, Zhiyang and Wang, Hongru and Cao, Zeyu and Dai, Jianbo and Wan, Yingjia and Guo, Zhijiang},
  booktitle={Advances in Neural Information Processing Systems},
  year={2024}
}

@article{DBLP:journals/jair/WeissI95,
  author       = {Sholom M. Weiss and
                  Nitin Indurkhya},
  title        = {Rule-based Machine Learning Methods for Functional Prediction},
  journal      = {J. Artif. Intell. Res.},
  volume       = {3},
  pages        = {383--403},
  year         = {1995},
  url          = {https://doi.org/10.1613/jair.199},
  doi          = {10.1613/JAIR.199},
  bibsource    = {dblp computer science bibliography, https://dblp.org}
}

@inproceedings{DBLP:conf/iclr/ZhangYMZY23,
  author       = {Shihao Zhang and
                  Linlin Yang and
                  Michael Bi Mi and
                  Xiaoxu Zheng and
                  Angela Yao},
  title        = {Improving Deep Regression with Ordinal Entropy},
  booktitle    = {The Eleventh International Conference on Learning Representations,
                  {ICLR} 2023, Kigali, Rwanda, May 1-5, 2023},
  publisher    = {OpenReview.net},
  year         = {2023},
  url          = {https://openreview.net/forum?id=raU07GpP0P},
  bibsource    = {dblp computer science bibliography, https://dblp.org}
}

@inproceedings{DBLP:conf/icml/ImaniW18,
  author       = {Ehsan Imani and
                  Martha White},
  editor       = {Jennifer G. Dy and
                  Andreas Krause},
  title        = {Improving Regression Performance with Distributional Losses},
  booktitle    = {Proceedings of the 35th International Conference on Machine Learning,
                  {ICML} 2018, Stockholmsm{\"{a}}ssan, Stockholm, Sweden, July
                  10-15, 2018},
  series       = {Proceedings of Machine Learning Research},
  volume       = {80},
  pages        = {2162--2171},
  publisher    = {{PMLR}},
  year         = {2018},
  url          = {http://proceedings.mlr.press/v80/imani18a.html},
  bibsource    = {dblp computer science bibliography, https://dblp.org}
}

@article{DBLP:journals/corr/abs-2110-14168,
  author       = {Karl Cobbe and
                  Vineet Kosaraju and
                  Mohammad Bavarian and
                  Mark Chen and
                  Heewoo Jun and
                  Lukasz Kaiser and
                  Matthias Plappert and
                  Jerry Tworek and
                  Jacob Hilton and
                  Reiichiro Nakano and
                  Christopher Hesse and
                  John Schulman},
  title        = {Training Verifiers to Solve Math Word Problems},
  journal      = {CoRR},
  volume       = {abs/2110.14168},
  year         = {2021},
  url          = {https://arxiv.org/abs/2110.14168},
  eprinttype    = {arXiv},
  eprint       = {2110.14168},
  bibsource    = {dblp computer science bibliography, https://dblp.org}
}

@article{DBLP:journals/tac/TsitsiklisR97,
  author       = {John N. Tsitsiklis and
                  Benjamin Van Roy},
  title        = {An analysis of temporal-difference learning with function approximation},
  journal      = {{IEEE} Trans. Autom. Control.},
  volume       = {42},
  number       = {5},
  pages        = {674--690},
  year         = {1997},
  url          = {https://doi.org/10.1109/9.580874},
  doi          = {10.1109/9.580874},
  bibsource    = {dblp computer science bibliography, https://dblp.org}
}

@article{DBLP:journals/corr/abs-2407-04811,
  author       = {Matteo Gallici and
                  Mattie Fellows and
                  Benjamin Ellis and
                  Bartomeu Pou and
                  Ivan Masmitja and
                  Jakob Nicolaus Foerster and
                  Mario Martin},
  title        = {Simplifying Deep Temporal Difference Learning},
  journal      = {CoRR},
  volume       = {abs/2407.04811},
  year         = {2024},
  url          = {https://doi.org/10.48550/arXiv.2407.04811},
  doi          = {10.48550/ARXIV.2407.04811},
  eprinttype    = {arXiv},
  eprint       = {2407.04811},
  bibsource    = {dblp computer science bibliography, https://dblp.org}
}

@article{DBLP:journals/corr/abs-2402-03300,
  author       = {Zhihong Shao and
                  Peiyi Wang and
                  Qihao Zhu and
                  Runxin Xu and
                  Junxiao Song and
                  Mingchuan Zhang and
                  Y. K. Li and
                  Y. Wu and
                  Daya Guo},
  title        = {DeepSeekMath: Pushing the Limits of Mathematical Reasoning in Open
                  Language Models},
  journal      = {CoRR},
  volume       = {abs/2402.03300},
  year         = {2024},
  url          = {https://doi.org/10.48550/arXiv.2402.03300},
  doi          = {10.48550/ARXIV.2402.03300},
  eprinttype    = {arXiv},
  eprint       = {2402.03300},
  bibsource    = {dblp computer science bibliography, https://dblp.org}
}

@article{yang2024qwen2,
  title={Qwen2 technical report},
  author={Yang, An and Yang, Baosong and Hui, Binyuan and Zheng, Bo and Yu, Bowen and Zhou, Chang and Li, Chengpeng and Li, Chengyuan and Liu, Dayiheng and Huang, Fei and others},
  journal={arXiv preprint arXiv:2407.10671},
  year={2024}
}
\bibliographystyle{icml2025}

%%%%%%%%%%%%%%%%%%%%%%%%%%%%%%%%%%%%%%%%%%%%%%%%%%%%%%%%%%%%%%%%%%%%%%%%%%%%%%%
%%%%%%%%%%%%%%%%%%%%%%%%%%%%%%%%%%%%%%%%%%%%%%%%%%%%%%%%%%%%%%%%%%%%%%%%%%%%%%%
% APPENDIX
%%%%%%%%%%%%%%%%%%%%%%%%%%%%%%%%%%%%%%%%%%%%%%%%%%%%%%%%%%%%%%%%%%%%%%%%%%%%%%%
%%%%%%%%%%%%%%%%%%%%%%%%%%%%%%%%%%%%%%%%%%%%%%%%%%%%%%%%%%%%%%%%%%%%%%%%%%%%%%%
\newpage
\appendix
\onecolumn
\section{Related Work}
\subsection{Test-time Scaling for LLM Reasoning.}
The test-time scaling technique requires models to generate long Chain-of-Thought(CoT) explicitly or implicitly as its thinking steps or reasoning steps, which can effectively improve the reasoning capabilities of large language models. Some works provide relative reasoning demonstrations or thinking paradigms on the input, using few-shot prompting techniques like CoT prompting\citep{DBLP:conf/nips/Wei0SBIXCLZ22} or in-context learning\citep{DBLP:journals/corr/abs-2211-09066} to achieve test-time scaling. These methods can improve the reasoning capabilities of large language models to some extent, but the performance is highly sensitive to the quality and quantity of few-shot demonstrations\citep{qin-etal-2024-context}. Some researchers focus on incorporating searching algorithm into the LLM reasoning scenario by explicitly performing searching algorithm during the reasoning process\citep{DBLP:conf/nips/YaoYZS00N23,DBLP:conf/emnlp/LiuG0HZQZ23,DBLP:conf/iclr/0002WSLCNCZ23,DBLP:conf/emnlp/ZhaoXKHX23}. The methods are shown to be effective at the cost of large token consumptions, and the human-crafted searching algorithm can only fit specific tasks\citep{DBLP:journals/corr/abs-2409-12183}, which makes the methods restricted. Another line of implementation of the test-time scaling technique is reinforcement learning. By providing suitable signals or feedback, the LLMs can learn to reason from scratch\cite{deepseekai2025deepseekr1incentivizingreasoningcapability} or the instruct-tuned checkpoint\cite{DBLP:conf/nips/ZelikmanWMG22}.

\subsection{Process-supervised verifier.}
Researchers have found that the process-supervised verifiers that are trained on fine-grained signals are effective for LLM reasoning and reinforcement learning, compared with outcome-supervised verifiers\cite{DBLP:conf/iclr/LightmanKBEBLLS24}. However, The definition of fine-grained signals is vague. \citet{DBLP:conf/iclr/LightmanKBEBLLS24} define the fine-grained signal as the stepwise calculate correctness and the signal is -1 if the reasoning step is incorrect, 1 if the reasoning step is correct, 0 if the reasoning step is neural. \citet{DBLP:conf/acl/WangLSXDLCWS24} define the fine-grained signal as the binary signal. Specifically, they map the Monte Carlo return to binary labels. If all outcomes are wrong, the signal of the current step is labeled to be 0. If any outcome is correct, the current step's signal is labeled 1. Some other works define the signal in a rule-based version, consisting of calculating error, formatting error, or else\citep{DBLP:conf/icml/XiCHJZHDLGWGSFZ24, DBLP:journals/corr/abs-2412-06559}. In the LLM reasoning scenario, as only the final outcome has specific and accurate labels, inductive bias is inevitably introduced when defining the fine-grained signal, which may result in reward hacking due to the definition flaw\citep{deepseekai2025deepseekr1incentivizingreasoningcapability}. Following previous works\citep{DBLP:journals/corr/abs-2406-06592,DBLP:journals/corr/abs-2408-03314,lu2024autopsv}, we define the fine-grained signal as the state value of the current step, which is calculated using the labels of final outcome only to reduce the potential human-crafted inductive bias.

\subsection{Regression as Classification.}
Several works have replaced regression with classification to improve performance in different domains\citep{DBLP:journals/jair/WeissI95,DBLP:conf/iclr/ZhangYMZY23}. Notably, \citet{DBLP:conf/icml/ImaniW18} proposed the HL-Gauss cross-entropy loss as the drop-in replacement of MSE loss, showing that by incorporating prior distribution information, the classification loss can be effective in regression tasks. The authors then performed experiments in several different distributions and found that the HL-Gaussian distribution can greatly improve the performance, while other distribution-learning approaches can result in even worse performance. \citet{DBLP:conf/icml/FarebrotherOVTC24} further extended the classification optimization method to several more complex domains, showing the general benefits of performing cross-entropy on the categorical distribution for regression tasks. However, how to perform better categorical distribution modeling is still unclear. In the LLM reasoning scenario, we endow the categorical distribution with fixed bins the specific definition, explaining the difference between regression and classification from the perspective of prior structural information injection. We show that the regression loss can also benefit from suitable categorical distribution modeling, indicating that the categorical distribution modeling can achieve the general benefits compared with scalar regression methods.

\section{More Implementation Details}
We provide the hyper-parameters when training the generator and value-based process verifiers in Table \ref{tab:hyperparams}.
\begin{table}[h!]
    \centering
    \begin{tabular}{c|c}
    \toprule
      Parameter   &  Value \\
    \midrule
      Epochs & 3(generator)/1(verifier) \\
      Learning Rate   &  2e-6 \\
      Batch size(per device) & 4(generator)/2(verifier) \\
      Gradient Accumulation Steps & 8 \\
      Max Sequence Length & 1024 \\
      Float Point Precision & torch.bfloat16 \\
      GPUs & 4 \\
    \bottomrule
    \end{tabular}
    \caption{The hyper-parameter when performing all the experiments.}
    \label{tab:hyperparams}
\end{table}

\section{Reasoning process transit from non-deterministic to deterministic.}\label{app:noise}
\begin{figure}[h!]
    \centering
    \includegraphics[width=0.4\textwidth]{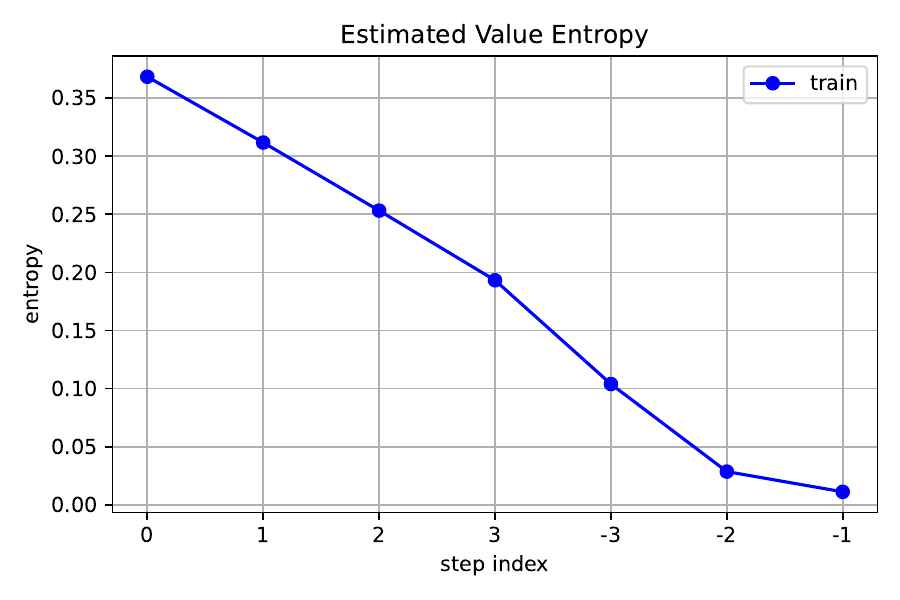}    
    \caption{The entropy of the estimated state values on the verifier's training dataset.}
    \label{fig:entropy}
\end{figure}

Finally, we further analyze the error and noise when estimating state value via Monte Carlo Sampling. First of all, we argue that the reasoning process transits from non-deterministic to deterministic. We measure the process as deterministic or non-deterministic by the entropy of the estimated state values as follows:
\begin{equation}
    E(k)=-\frac{1}{N}\sum_i^N H(\hat{V}^\pi(s_i^k))+H(1-\hat{V}^\pi(s_i^k)), \nonumber
\end{equation}
and
\begin{equation}
    H(\hat{V}^\pi(s_i^k))=\hat{V}^\pi(s_i^k)\log \hat{V}^\pi(s_i^k), \nonumber
\end{equation}
where $s_i^k$ is the $k$-th state of the $i$-th data. The entropy will be close to 0 when the estimated value at the $k$-th step is close to marginal values(i.e. 0 or 1), and will achieve its maximum when the estimated value at the $k$-th step is close to the middle(i.e. 0.5). We estimate the entropy by our verifier's training dataset, created by the fine-tuned LLemma-7b. The average steps of our training dataset is 7.8, so we tallied the results of the first 4 steps and the last 3 steps, labeled as \{1, 2, 3, 4, -3, -2, -1\}. The results are shown in Figure\ref{fig:entropy}. As shown in the figure, the entropy of the estimated state value keeps decreasing as the reasoning process goes, which means that the estimated state value as well as the ground-truth probability $p$ is getting close to marginal, assuming that the estimated state value can be the proxy to the ground-truth probability $p$. By estimating the ground-truth categorical distribution with Gaussian distribution $X\sim N(p,\frac{p(1-p)}{k})$ and let $k=8$, the standard deviation of $X$ when $p$ is 0.5 is about 0.177. As the bin width of the categorical distribution is $\frac{1}{k}=0.125$, the three-sigma region almost covers the whole categorical distribution, which shows that the one-time sampling from the categorical distribution to estimate the state value induces non-negligible errors, especially in early steps of the whole reasoning process.

%%%%%%%%%%%%%%%%%%%%%%%%%%%%%%%%%%%%%%%%%%%%%%%%%%%%%%%%%%%%%%%%%%%%%%%%%%%%%%%
%%%%%%%%%%%%%%%%%%%%%%%%%%%%%%%%%%%%%%%%%%%%%%%%%%%%%%%%%%%%%%%%%%%%%%%%%%%%%%%

\end{document}